\documentclass[english]{article}
\usepackage[T1]{fontenc}
\usepackage[utf8]{inputenc}
\usepackage{graphicx}
\usepackage{hyperref}

\makeatletter

\providecommand{\tabularnewline}{\\}

\makeatother

\usepackage{babel}
\begin{document}
\title{An ontology for the formalization and visualization of scientific
knowledge\thanks{This paper is as an extended, translated, version of \cite{dapfal2018} }}
\author{Vincenzo Daponte, Gilles Falquet}
\date{Centre Universitaire d'informatique, \\
Universit\'e de Gen\`eve, Switzerland\\
\texttt{\small{}Vincenzo.Daponte@unige.ch, Gilles.Falquet@unige.ch}}
\maketitle
\begin{abstract}
The construction of an ontology of scientific knowledge objects, presented
here, is part of the development of an approach oriented towards the
visualization of scientific knowledge. It is motivated by the fact
that the concepts that are used to organize scientific knowledge (theorem,
law, experience, proof, etc.) appear in existing ontologies but that
none of these ontologies is centered on this topic and presents them in a simple and easily
understandable organization. This ontology has been constructed by 1) selecting concepts that appear in high level ontologies or in ontologies of knowledge objects of specific fields and 2)  interviewing scientists in different fields. We have aligned this ontology with some of the sources
used, which has allowed us to verify its consistency with respect
to them. The validation of the ontology consists in using it to formalize
knowledge from various sources, which we have begun to do in the field
of physics.

\textbf{keywords: }Ontology, Scientific knowledge, Knowledge visualization
\end{abstract}

\section{Motivations}

The access to scientific knowledge, whether general or factual, must
necessarily involve a visual, auditory or other presentation that
appeals to one or more senses of the human being. If we are interested
in the visual presentation of knowledge, we notice that the natural
written language occupies a predominant place in it but that other
graphic forms (notations, mathematical and chemical formulas, diagrams,
tables, forms, hypertexts, etc.) play an important role in facilitating
the performance of various intellectual tasks (calculation, comparison,
deduction, etc.). 

The general framework in which our work is carried out is the study
of techniques for visualizing scientific knowledge and in particular
their formal specification with the purpose of building visualization
tools that are adapted to the tasks of the scientific user. Indeed,
experience shows that there is not an optimal visualization technique
but that the effectiveness of each technique depends on the context
and objectives of the user (see, for example \cite{Card1999}).

\begin{figure}
\begin{centering}
\includegraphics[width=1\columnwidth]{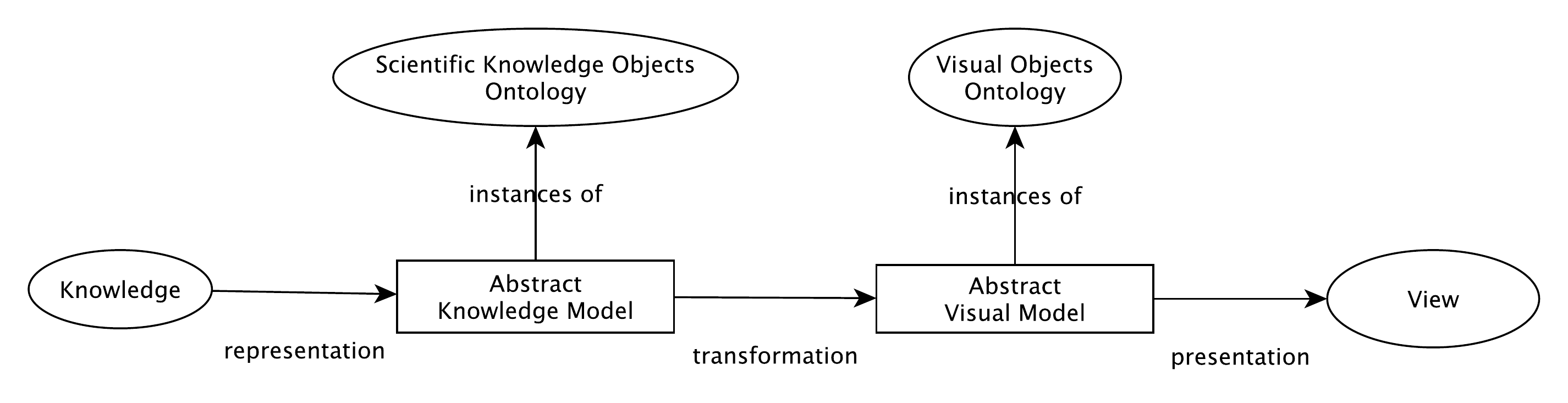}
\par\end{centering}
\caption{\label{fig-viz}Adaptation of the \cite{Chi2000} model to the visualization
of scientific knowledge }
\end{figure}

To formally represent the notion of visualization technique it is
necessary, according to the reference model proposed by \cite{Chi2000},
To formally represent the notion of visualization technique, it is
necessary to define an abstract model of the data to be visualized,
an abstract model of the visual objects and an application of the
data model in the abstract visual model. In the case of the visualization
of scientific knowledge, it is therefore necessary to create an abstract
model of the scientific knowledge to be visualized. Our goal is to
provide a formalization of visualization that applies to any science,
so we decided to build an ontology of the knowledge structuring objects
used in the various sciences. Such an ontology will make it possible
to model the knowledge to be visualized in the form of instances of
the classes of this ontology, according to the diagram in figure \ref{fig-viz}.

If sciences vary according to their objects of study (differentiable
functions, butterflies, human societies, elementary particles,...),
they also have their own concepts to structure the knowledge produced.
Mathematics produces theorems, corollaries, lemmas, conjectures, proofs,
whereas physics speaks about laws, principles, measures, experimental
results or anthropology produces observations, theories, explanatory
hypotheses, methods. In addition, each science has developed a set
of formalisms (i.e. linguistic constructs) to express and process
this knowledge: chemical formulae and equations, mathematical formulae,
flow diagrams, interaction diagrams, syntactic trees, etc. and a set
of knowledge production techniques: experimentation, formal reasoning,
surveys, observations, etc. However, it can be seen that the vocabulary
for classifying scientific knowledge objects is limited and that several
terms often cover similar concepts. Therefore, we can think of building
a central ontology composed of a small number of classes of scientific
objects. The knowledge to be visualized can then be represented as
instances of these classes. 

In the rest of this article we will begin by examining the work that
focuses on the representation of scientific knowledge and those that,
while somewhat related, can provide important elements. We will then
present the method used to create a first version of the SKOO ontology
of scientific knowledge objects and the ontology obtained. We will
then present the first evaluations we have carried out. In the conclusion
we will give perspectives on the practical use of this ontology and
on the continuation of the evaluation and development of the ontology.

\section{State of the art}

To our knowledge, there is currently no ontology whose field is the
conceptualization of objects used to represent or structure scientific
knowledge in general. It should also be noted that works on the epistemology
of science do not generally address this general ontological question
but deal either with a particular science or a particular aspect of
science. On the other hand, this ontological work was carried out
in knowledge engineering for some specific fields. 

The OMDoc ontology core presented in (\cite{lange2013ontologies})
presents a model of mathematical knowledge in the form of knowledge
items that are types of mathematical objects, theories or statements,
which can be of the \emph{Assertion}, \emph{Proof}, \emph{Definition},
\emph{Axion}, etc. type. In addition, there are relationships between
these types of elements, such as the \emph{Proof}-\emph{Assertion}
proves relationship. Although this ontology is dedicated to mathematics,
it can easily be extended to other sciences.

The SIO (Semanticscience Integrated Ontology) ontology, \cite{Dumontier2014},
is a higher level ontology that essentially aims to represent biomedical
knowledge. The primary purpose of SIO is to describe complex biomedical
objects (with component-composite relationships) and the processes
in which they are involved or the (experimental) procedures applied
to them. However, in the SIO description class, there are many concepts
used to structure scientific knowledge: argument, belief, conclusion,
evidence, hypothesis,... However, unlike OMDoc, there are no specific
relationships between these concepts, but they can be linked to the
objects they describe. SIO also describes the linguistic, mathematical
and media objects that support knowledge.

The notion of experimental scientific process is at the heart of the
EXPO system (\cite{soldatova2006ontology}). It combines SUMO ontology
(\cite[)]{niles2001towards}) with subject-specific ontologies of
experiments by formalizing the generic concepts of experimental design,
methodology and results representation. EXPO aims to describe different
experimental domains and to provide a formal description of the experiences
for analysis, annotation and sharing of results.

We can also consider what has been done in scientific knowledge bases,
such as Gene Ontology (\cite{ashburner2000gene}), OntoMathPro (\cite{nevzorova2014ontomath})
or FMA (\cite{Rosse2007}) that aim to represent the current state
of our knowledge in a field. They generally consist of a terminology
part that organizes very precisely the concepts of the domain and
a part composed of assertions (statements) that represent our knowledge
about these concepts. In Gene Ontology (GO), statements are called
\textquotedbl annotations\textquotedbl . They typically link a gene
and a term from the GO ontology (for example, to indicate that the
gene has a certain function). The statements are qualified by a type
of proof (experimental, phytogenetic inference, automatic inference,
etc.). In OntoMathPro (\cite{nevzorova2014ontomath}) the terminology
levels and assertions do exist but are not structurally separated.
Thus Stokes' theorem (assertion) is not an instance but a subclass
of the \emph{Theorem} class. Similarly, as a first level subclass
of Mathematical knowledge object we find both Theorem and Tensor.
There is therefore an aggregation of the objects describing the knowledge
and the objects of the domain on which we are working.

It should also be noted that there are ontologies whose sole purpose
is to list and classify the subject-specific objects or to create
a domain terminology (SWEET, ScienceWISE, etc.). In general, these
ontologies are not interested in knowledge structuring objects. 

On the other hand, a lexical ontology such as WordNet contains a large
number of concepts such as theorem, law, definition, hypothesis, corollary.
However, it should be noted that these concepts are not organized
in a way that can be directly used. For example, we have the hyperonymic
relationship chains

\[
theorem<idea<content<cognition
\]
 and 

\emph{
\[
corollary<...<process<content<cognition
\]
}

\noindent whereas from a formal point of view a corollary is a theorem.
In other words, an ontology of scientific knowledge objects cannot
be extracted by the simple projection of a part of WordNet. The same
is true for other higher-level ontologies (SUMO, CyC,...)

\section{Construction of the SKOO ontology}

\begin{figure}[h]
\begin{centering}
\includegraphics[width=1\columnwidth]{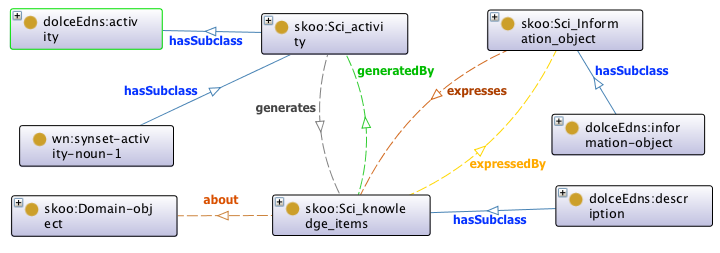}\caption{\label{Fig.2.niveau-sup-ontologie}Structure of the top level of the
SKOO ontology.}
\par\end{centering}
\end{figure}

To build the ontology of \emph{Scientific Knowledge Objects Ontology}
(SKOO)\footnote{\url{http://purl.org/net/skoo}} we applied the following process:
\begin{enumerate}
\item We collected a set of terms used to structure knowledge in different
scientific fields. This was done by consulting books (textbooks, forms,
``handbooks'' monographs) in biochemistry, physics, mathematics,
linguistics, sociology; interviews with scientists from different
fields; analysis of the terminology level of scientific knowledge
bases and ontologies (Gene Ontology, OntoMathPro,...)
\item To build the upper level of ontology we first associated the highlighted
terms with equivalent or more general ``synsets'' of WordNet. Then
we used the DOLCE ontology, already aligned with WordNet, to find
higher-level concepts.
\item Finally, we have defined relationships between higher-level concepts
based on relationships found in scientific ontologies, in particular
OMDoc, and by specialization of certain high-level DOLCE relationships.
\end{enumerate}
Figure \ref{Fig.2.niveau-sup-ontologie} shows the upper level of
the ontology obtained and its links with DOLCE and WordNet. We describe
below the interpretation of each of its classes.

\begin{description}
\item [{Sci\_Knowledge\_Item}] The items of scientific knowledge are all
the objects that serve to structure the expression of scientific knowledge.
They may be objects, such as theorems, laws (physical, chemical),
models or methods that carry knowledge in themselves in the Platonic
sense of true and justified belief. But they can also be ``auxiliary''
objects such as definitions, examples, evidence, hypotheses, problems.
These objects correspond to the objects of the \emph{description}
class of the ontology DOLCE (\cite{masolo2003wonderweb}). 
\item [{Sci\_Information\_Object}] The purpose of this class is to group
all forms of expression of knowledge elements, whether linguistic
or in the form of diagrams, schemas, formulas, etc.. This is a subclass
of the DOLCE ontology \emph{information-object} class (\cite{masolo2003wonderweb}),
and its main class \emph{Sci-linguistic-object} is a subclass of the
DOLCE \emph{linguistic-object} class (\cite{masolo2003wonderweb}).
This class aims to include all forms and methods of expressing the
concepts used to represent knowledge of the disciplines under consideration. 
\item [{Sci\_Activity}] This class represents the activities, in the sense
of \emph{activity} (hyponym of \emph{human activity}) in WordNet,
that are used to generate elements of scientific knowledge. These
activities may be experimental (process, experimentation, observation),
but also empirical (conducting surveys) or formal (formally proving,
calculating). The precise description of activities, in particular
experimentation, is not defined in this ontology because it is already
covered by other ontologies, such as SIO and EXPO. 
\item [{\emph{Domain-object}}] represents all objects about which scientific
knowledge is expressed. This class serves as an anchor point for classes
describing the objects studied in specific fields. When using ontology
in practice, the principle is to import an ontology of objects from
the scientific domain concerned and create subsumption axioms $C\sqsubseteq Domain-object$
for its higher level classes.
\end{description}
\begin{figure}[h]
\begin{centering}
\includegraphics[scale=0.5]{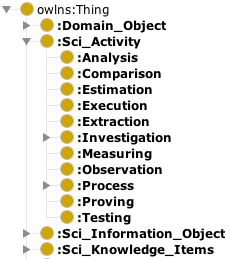}\includegraphics[scale=0.5]{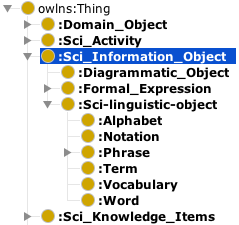}\includegraphics[scale=0.5]{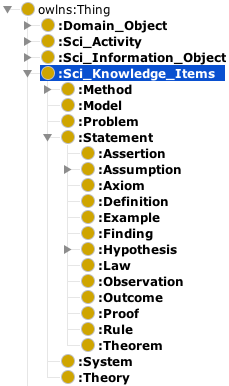}
\par\end{centering}
\caption{\label{Fig.3}The higher levels of the subclasses of \emph{Sci-activity},
\emph{Sci-knowledge-items }and \emph{Sci-information-object.} }
\end{figure}

\section{Evaluation}

We conducted two types of evaluations, consistency and capacity. In
addition to the internal consistency of the ontology, to give an indication
of the external consistency (relative to other ontologies), the ontology
has been aligned with the OMDoc, DOLCE and WordNet ontologies. To
do this, we translated the OMDoc and WordNet concepts into OWL classes,
then created correspondence axioms of the owl:subClassOf and owl:EquivalentClass
types between them and SKOO. Table \ref{Tab.1}  shows some of these
axioms. We then verified the consistency of the ontologies obtained
by merging SKOO, the three ontologies and the correspondence axioms
(but without correspondence between DOLCE, WordNet and OMDoc).

\begin{table}
\begin{centering}
\begin{tabular}{|c|c|c|c|}
\hline 
{\footnotesize{}SKOO} & {\footnotesize{}OMDoc} & {\footnotesize{}DOLCE} & WordNet\tabularnewline
\hline 
\hline 
{\footnotesize{}Sci-knowledge-items} & {\footnotesize{}$\sqsupseteq$ MathKnowledgeItem} & {\footnotesize{}$\sqsubseteq$ description} & \tabularnewline
\hline 
{\footnotesize{}Statement} & {\footnotesize{}$\sqsupseteq$ Statement} &  & {\footnotesize{}$\sqsubseteq$ statement}\tabularnewline
\hline 
{\footnotesize{}Theory} & {\footnotesize{}$\sqsupseteq$ Theory} & {\footnotesize{}$\sqsubseteq$ theory} & {\footnotesize{}$\sqsubseteq$ theory}\tabularnewline
\hline 
{\footnotesize{}Assertion} & {\footnotesize{}$\sqsupseteq$ Assertion} &  & {\footnotesize{}$\sqsubseteq$assertion}\tabularnewline
\hline 
{\footnotesize{}Axiom} & {\footnotesize{}$\sqsupseteq$ Axiom} &  & {\footnotesize{}$\sqsubseteq$axiom}\tabularnewline
\hline 
{\footnotesize{}Definition} & {\footnotesize{}$\sqsupseteq$ Definition} &  & {\footnotesize{}$\sqsubseteq$definition}\tabularnewline
\hline 
{\footnotesize{}Proof} & {\footnotesize{}$\sqsupseteq$ Proof} &  & {\footnotesize{}$\sqsubseteq$ proof}\tabularnewline
\hline 
{\footnotesize{}Sci-activity} &  &  & {\footnotesize{}$\sqsubseteq$ activity}\tabularnewline
\hline 
{\footnotesize{}Process} &  & {\footnotesize{}$\sqsupseteq$ activity} & {\footnotesize{}$\sqsubseteq$ process}\tabularnewline
\hline 
{\footnotesize{}Sci-information-object} &  & {\footnotesize{}$\sqsubseteq$ information-object} & \tabularnewline
\hline 
\end{tabular}
\par\end{centering}
\caption{\label{Tab.1}Correspondence between SKOO and OMDoc, DOLCE, WordNet
classes.}
\end{table}

\begin{figure}
\begin{centering}
\includegraphics[width=1\columnwidth]{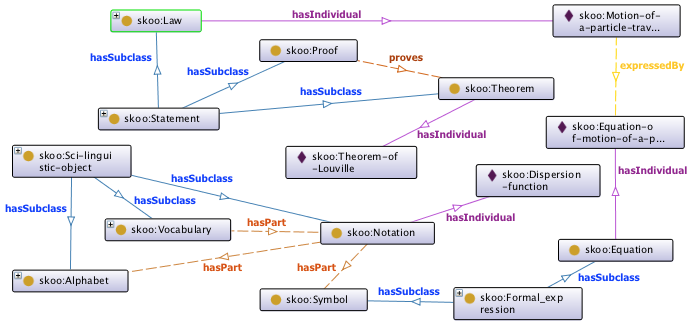}
\par\end{centering}
\caption{\label{Fig.4.skoo.physics}Some concepts contained in several sections
of Chapter 3 of (\cite{wille2000physics}) expressed as instances
of SKOO ontology.}

\end{figure}

To evaluate the capabilities of ontology we must verify whether, given
a system of visualization of scientific knowledge, ontology makes
it possible to create an appropriate abstract model for this knowledge.
In the case of structured and homogeneous knowledge bases, such as
Gene Ontology annotations or mathematical forms, it is easy to check
the adequacy of the ontology. Indeed, this knowledge generally corresponds
to \emph{statements} that can be theorems. On the other hand, the
case of knowledge expressed in texts is more complex. We carried out
a first test by taking as a visualization system a part of an accelerator
physics book (\cite{wille2000physics}). We have modeled different
concepts from the different sections of Chapter 3, in particular,
the main concept expressed in Section 3.2, the complete Section 3.6
of this book (\emph{dispersion and momentum compaction factor}) and
a theorem used in Section 3.8 as SKOO class instances. Figure \ref{Fig.4.skoo.physics}
shows the modeling of a physical law (\emph{Law} instance) represented
by an equation (\emph{Equation} instance) and also the modeling of
a theorem (\emph{Theorem} instance) and a notation (\emph{Notation}
instance) used to represent a particular concept. It should be noted
that in the relationships shown in this figure, the \emph{hasIndividual}
relationship is used by \emph{Protégé} to associate the type of instance
(the class) with the instance itself.

\section{Conclusions and future work}

We presented the construction of the first version of the SKOO ontology
whose purpose is to provide a general model for the modeling of scientific
knowledge to be visualized (according to the diagram of \cite{Chi2000}).
Although the concepts represented in this ontology all exist in other
ontologies, none of them are grouped in such a way that they can be
directly used to represent scientific knowledge. Hence the interest
of the SKOO ontology. The ontology was aligned with reference ontologies
to verify with a reasoner that they did not contradict them. On the
other hand, we have begun to validate the capacity of this ontology
to model the knowledge represented in existing knowledge bases, which
does not pose any particular problem, and the knowledge represented
in (hyper)texts, which is more difficult, especially for texts in
the human and social sciences. After having carried out a test on
a part of a physics book, we will undertake tests on books from the
humanities and social sciences.

The next step in this work will be to fully model various existing
knowledge visualization systems. To do this, we will use the SPARQL
language to specify transformations from a knowledge model (expressed
with SKOO) to a model of visualization objects (including lists, trees,
graphs, texts, geometric shapes, etc.). This will validate the complete
model for specifying visualization techniques. From there it will
be possible to create a visualization generation system from their
specification and use it to create new visualization techniques and
test them with users.

During this work we realized that the interest of this ontology goes
beyond the mere visualization of knowledge. It is applicable, for
example, in the context of the search for precise information or automatic
reasoning on large bodies of scientific knowledge.

\paragraph{Acknowledgement}

This work has been enriched by the valuable advice and experience
of Giuseppe Cosenza and Jean-Pierre Hurni, whom the authors would
like to thank.

\bibliographystyle{plain}
\bibliography{skoo-ic2018-eng}

\end{document}